\pdfoutput=1
\documentclass[11pt]{article}
\usepackage[preprint]{acl}

\usepackage{times}
\usepackage{latexsym}
\usepackage[T1]{fontenc}
\usepackage[utf8]{inputenc}
\usepackage{microtype}
\usepackage{inconsolata}
\usepackage{graphicx}
\usepackage[commandnameprefix=ifneeded]{changes}
\definechangesauthor[color=magenta]{af}
\definechangesauthor[color=orange]{sa}
\definechangesauthor[color=blue]{am}
\definechangesauthor[color=olive]{ng}
\definechangesauthor[color=olive]{tacl}

\usepackage{booktabs}
\usepackage{multirow}

\usepackage{arydshln}
\usepackage{makecell}
\makeatletter
\def\adl@drawiv#1#2#3{%
        \hskip.5\tabcolsep
        \xleaders#3{#2.5\@tempdimb #1{1}#2.5\@tempdimb}%
                #2\z@ plus1fil minus1fil\relax
        \hskip.5\tabcolsep}
\newcommand{\cdashlinelr}[1]{%
  \noalign{\vskip 1.3pt
           \global\let\@dashdrawstore\adl@draw
           \global\let\adl@draw\adl@drawiv}
  \cdashline{#1}[.4pt/2pt]
  \noalign{\global\let\adl@draw\@dashdrawstore
           \vskip 1.3pt}}
\makeatother

\usepackage{amsmath}
\usepackage[capitalize,nameinlink]{cleveref}
\usepackage{hyperref}
\usepackage{tikz}
\definecolor{tikz_blue}{HTML}{009ADE}
\definecolor{tikz_red}{HTML}{F67280}
\definecolor{tikz_orange}{HTML}{F28522}
\definecolor{tikz_gray}{HTML}{B3B3B3}
\definecolor{tikz_green}{HTML}{91B54D}
\usepackage{pgf-pie}
\usepackage{pgfplots}
\usetikzlibrary{pgfplots.groupplots}
\pgfplotsset{compat=1.3}
\usepackage{tikz}
\usetikzlibrary{patterns}
\usetikzlibrary{shapes.geometric}
\usepackage{colortbl}
\usepackage{xcolor}
\Crefname{section}{\S\hspace{-1mm}}{\S\hspace{-0.5mm}}
\Crefname{appendix}{App.}{Apps.}

\title{Translate Smart, not Hard:\\Cascaded Translation Systems with Quality-Aware Deferral}

\author{António Farinhas$^{1,2}$, Nuno M. Guerreiro$^{1,2,3,4}$, Sweta Agrawal$^{1}$, \\
\textbf{Ricardo Rei}$^{4}$\textbf{, André F.T. Martins}$^{1,2,4,5}$ \\
$^1$Instituto de Telecomunicações, $^2$Instituto Superior Técnico, Universidade de Lisboa
\\$^3$MICS, CentraleSupélec, Université Paris-Saclay, $^4$Unbabel, $^5$ELLIS Unit Lisbon\\
\small\texttt{antonio.farinhas@tecnico.ulisboa.pt}}

\begin{document}
\maketitle
\begin{abstract}
Larger models often outperform smaller ones but come with high computational costs.
Cascading offers a potential solution. By default, it uses smaller models and defers only some instances to larger, more powerful models.
However, designing effective deferral rules remains a challenge.
In this paper, we propose a simple yet effective approach for machine translation, using existing quality estimation (QE) metrics as deferral rules. We show that QE-based deferral allows a cascaded system to match the performance of a larger model while invoking it for a small fraction ($30\%$ to $50\%$) of the examples, significantly reducing computational costs. We validate this approach through both automatic and human evaluation.
\end{abstract}

\section{Introduction}

Larger models consistently outperform smaller ones in NLP tasks, but the trade-off is the increased computational cost.
This raises the question: 
\begin{quote}
    \textit{How can we maintain high performance while reducing computational load?}
\end{quote}
A promising solution is \textbf{model cascading}, where smaller models handle examples by default, and only a subset of hard instances is deferred to a larger model.
However, this approach requires a robust deferral system that reliably determines when to defer. 
Common approaches often involve designing and training specialized deferral models, which determine when a large model is needed---\textit{e.g.}, based on reliability or uncertainty estimates \citep{chen2023frugalgptuselargelanguage, gupta2024language}.
But do we really need to train new models for every task, or can existing resources speed up this process?

\begin{figure}[t]
    \centering
    \includegraphics[width=0.99\linewidth]{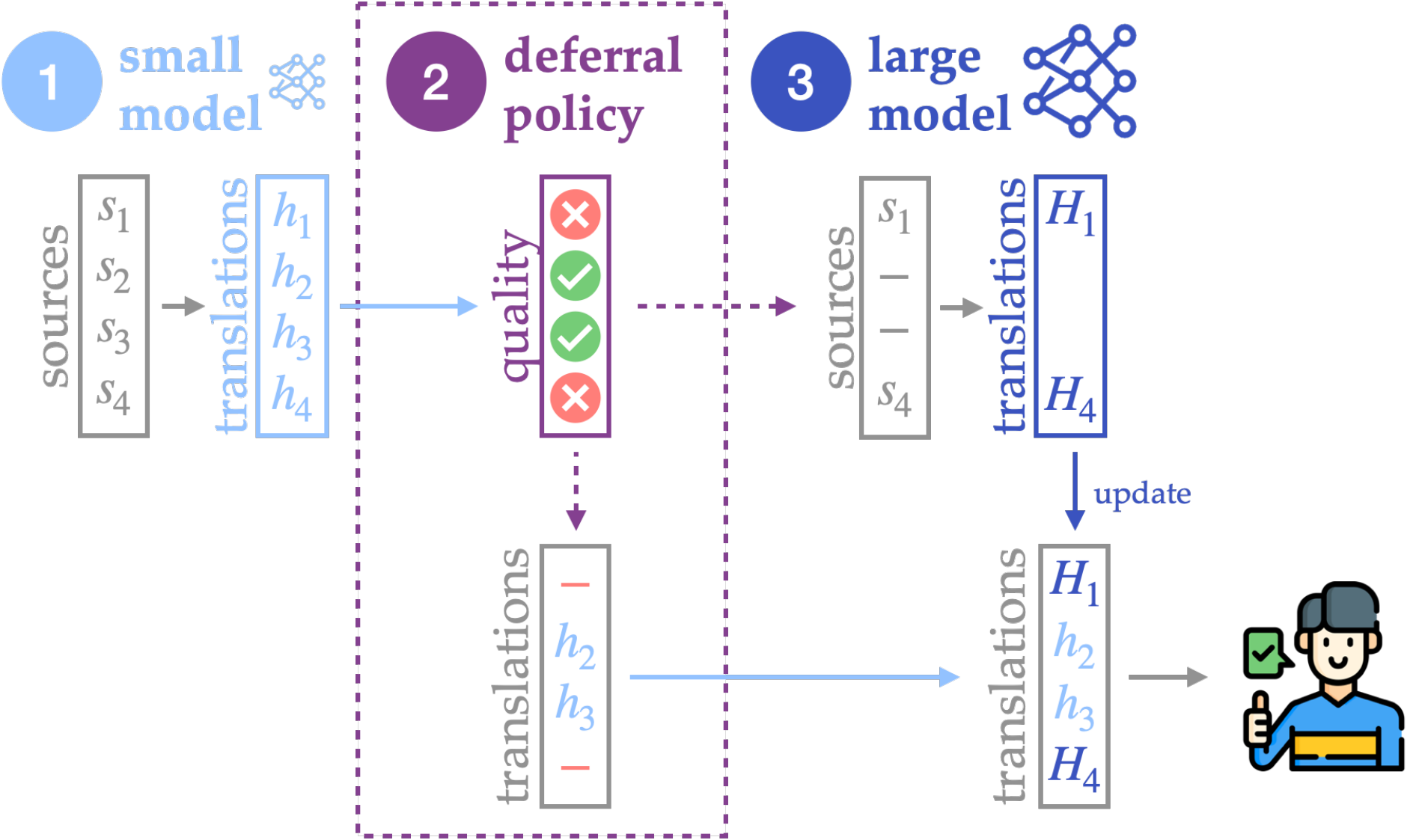}
    \caption{Cascaded translation system with QE-based deferral. A small model translates a batch of source sentences, and a relatively lightweight QE model scores the hypotheses. Sources with the lowest-scoring translations are deferred to a larger model. The extent of deferral is determined by a predefined compute budget.}
    \label{fig:cascading-system}
\end{figure}

For \textbf{machine translation} (MT), extensive research on reference-free automatic evaluation offers an appealing alternative \citep{zerva-etal-2022-findings, zerva-etal-2024-findings, blain-etal-2023-findings}.
In this paper, we leverage recent \textbf{quality estimation} (QE) metrics to create straightforward and relatively lightweight deferral rules. 
This approach draws inspiration from professional translation workflows, where QE metrics help identify translations that should be deferred to expert post-editing \citep{castilho2017acceptability, bechara2021the}.
Our main contributions are:

\begin{itemize}
    \item We introduce a \textbf{cascaded translation system} that uses pretrained QE metrics to determine whether to defer examples from a smaller model to a larger one, balancing efficiency and quality (\cref{sec:An Embarrassingly Easy Approach for Machine Translation}).
    See \cref{fig:cascading-system} for an illustration.
    \item We confirm that the benefits of QE-based model cascading hold across different combinations of translation and QE models (\cref{sec:Experiments and Analysis}).
    \item We perform \textbf{human evaluation}, further validating our approach on two language pairs (en-es and en-ja) in the WMT24 test set (\cref{sec:Human Evaluation}).
    \item We release our code, all generated translations, and human quality assessments.\footnote{All resources are available at: \url{https://github.com/deep-spin/translate-smart}.}
\end{itemize}

\section{Adaptive Inference in NLP}
\label{sec:Adaptive Inference in NLP}

\textbf{Adaptive inference} techniques are increasingly being adopted in natural language processing tasks \citep{mamou2022tangobertreducinginferencecost, varshney-baral-2022-model, chen2023frugalgptuselargelanguage, ong2024routellmlearningroutellms}.
These methods typically use models of different sizes and predictive power (often two, though most frameworks can easily accommodate more), with the primary goal of reducing the computational load by using the larger, more computationally expensive model only when necessary (\textit{e.g.}, for more difficult examples or when a model is highly uncertain about its prediction).
Current strategies include \textbf{routing}, where a decision rule determines which model to use, ensuring only one model is used to handle each input, and \textbf{cascading}, which starts with a smaller model and may invoke a larger one afterward based on the small model's output and a deferral rule.
In this paper, we focus on the second approach.

The computational efficiency of model cascading comes at the cost of designing a \textbf{robust deferral system} that can reliably identify when to defer to the larger model. This is often handled using simple decision rules, such as nonparametric methods or other approaches based on uncertainty measures \citep{ramírez2024optimisingcallslargelanguage, gupta2024language}.
A recent alternative involves training external models specifically to predict when deferral is needed -- for a given example, these models can be trained, \textit{e.g.}, to assess if a given candidate is correct \citep{chen2023frugalgptuselargelanguage}.\footnote{Likewise, routing typically involves training external models to \textit{(i)} predict the performance of the small model \citep{sakota2024fly-swat}, or \textit{(ii)} determine if the small model is likely to outperform the large one \citep{ding2024hybrid}.}
Here, we propose a simple and effective deferral rule for MT that is conceptually similar to this approach while offering a particularly straightforward solution for this task.

\section{Quality-Aware Deferral for MT}
\label{sec:An Embarrassingly Easy Approach for Machine Translation}

Although human evaluations and reference-based metrics remain the standard for evaluating machine translations, reference-free/quality estimation (QE) metrics have shown strong correlations with human judgments \citep{zerva-etal-2024-findings}, holding promise in distinguishing between the quality of translations for the same source \cite{agrawal-etal-2024-automatic-metrics}.
Since QE models are typically much smaller than current translation models \citep{kocmi-etal-2024-findings},
we propose to leverage them for an efficient deferral rule.
Rather than training new bespoke decision models (\cref{sec:Adaptive Inference in NLP}), existing QE models can evaluate translations from a lightweight model and determine when to accept them or defer to a larger one.

\paragraph{How to choose which examples to defer?}
Setting a fixed threshold on QE scores is challenging---too high a threshold wastes computational resources, while too low a threshold risks compromising quality.
Throughout this paper, we use a \textbf{budget-constrained computation} approach: we first translate all examples in a batch with the smaller model, then rank them based on QE scores, deferring only the lowest-scoring subset according to a predefined compute budget (the fraction of examples deferred to the larger model).
This assumes parallel processing of entire batches rather than processing individual instances sequentially.
We leave alternatives such as dynamic thresholding \citep{ramírez2024optimisingcallslargelanguage} for future work.
See \cref{fig:cascading-system} for an illustration with $50\%$ of deferral.

\paragraph{Computational efficiency.}
The standard approximation for the number of floating point operations (FLOPs) required for inference with a transformer model is $2ND$, where $N$ represents the number of model parameters and $D$ is the number of tokens generated at inference time \citep{sardana2024beyond, snell2024scaling}.
For a cascaded approach with superscripts $S$ and $L$ denoting the smaller and larger models, respectively, this becomes:
\begin{equation}
    2BD_S(N_S + N_{QE}) + 2\eta BD_LN_L,
\end{equation}
where $B$ is the batch size and $\eta$ is the proportion of instances the larger model processes. Assuming $D_S\approx D_L$, this approach achieves computational parity with the larger model (\textit{i.e.}, $2BDN_L$) when:
\begin{equation}
    \eta^\star=1-\frac{N_S+N_{QE}}{N_L}.\label{eq:eta}
\end{equation}
This expression provides a simple rule of thumb: to maintain computational efficiency, the larger model should handle at most $\eta^\star$ of the examples. For instance, if it is $10\times$ larger than the smaller model and the QE model is negligible ($N_{QE}\ll N_S$), then $\eta^\star\approx0.9$. This means the cascading is more efficient than always using the larger model as long as fewer than $90\%$ of the examples are deferred.

\section{Experiments and Analysis}
\label{sec:Experiments and Analysis}

We consider Tower-v2 models \citep{rei-etal-2024-tower} of different size and predictive power: \textbf{Tower-v2 70B}, an improved iteration of Tower \citep{alves2024tower}, obtained by continued pretraining Llama-3 \citep{llama3modelcard} on a multilingual dataset with 25 billions of tokens, followed by supervised finetuning for translation-related tasks;\footnote{Combined with quality-aware decoding \citep{fernandes-etal-2022-quality}, this is the winning submission of the WMT24 general translation shared task \citep{kocmi-etal-2024-findings}.} and \textbf{Tower-v2 7B}, a more lightweight version using Mistral \citep{jiang2023mistral7b}.
Check \cref{app:Implementation details} for more details.

\paragraph{Deferral.}
We use two versions of \textsc{CometKiwi}:
\texttt{wmt22-cometkiwi-da} \citep{rei-etal-2022-cometkiwi}, which with only 0.5B parameters achieves a strong correlation with human judgments \citep{zerva-etal-2022-findings}; and \texttt{wmt23-cometkiwi-da-xxl} \citep{rei-etal-2023-scaling}, a scaled version with 10.5B parameters.
As baselines, we consider \texttt{random} selection; deferral rules based on source length computed using Tower-v2's tokenizer, \textit{i.e.}, deferring either the shortest (\texttt{length}) or the longest (\texttt{-length}) sources;\footnote{Source length is often used to assess translation difficulty \cite{kocmi-bojar-2017-curriculum, wan-etal-2022-challenges, wang-etal-2023-document-level}.} and a confidence measure based on the smaller model’s normalized log-probability (\texttt{logprobs}), \textit{i.e.}, deferring texts with the lowest likelihoods.
We also compare our approach with quality-aware decoding \citep{fernandes-etal-2022-quality} in \cref{App:Quality-Aware Decoding}.

\paragraph{Evaluation.}
We use the WMT24 test sets \citep{kocmi-etal-2024-findings}, which span multiple domains (news, social, speech, and literary) and $11$ language pairs (en-cs, en-de, en-es, en-hi, en-is, en-ja, en-ru, en-uk, en-zh, cs-uk, and ja-zh). For each language pair, we treat the full test set as a single batch for computing QE thresholds (\cref{sec:An Embarrassingly Easy Approach for Machine Translation}).\footnote{Results are then averaged across language pairs for better visualization unless otherwise stated.}
We evaluate systems with \textsc{metricX} \citep{juraska-etal-2023-metricx} to reduce the risk of ``reward hacking'' \citep{fernandes-etal-2022-quality} and better reflect real quality improvements.
Since biases may still exist when using a different evaluation metric than the reward model \citep{kovacs-etal-2024-mitigating}, we also conduct human evaluation (\cref{sec:Human Evaluation}).

\begin{table}[t]
\small
\centering
\setlength\tabcolsep{4pt}
\begin{tabular}{l ccc c}
\toprule
& M $\uparrow$ & C $\uparrow$ & Win rate \\
\midrule
Tower-v2 7B
& -3.01 & 83.94 & 
\scriptsize{\textcolor{tikz_red}{43\%}}
\begin{tikzpicture}
\draw [fill=tikz_red] (0,0) rectangle (0.43*1.4, 0.2);
\draw [fill=tikz_gray] (0.43*1.4,0) rectangle (0.68*1.4, 0.2); 
\draw [fill=tikz_green] (0.68*1.4,0) rectangle (1*1.4, 0.2); 
\end{tikzpicture} \scriptsize{\color{tikz_green}32\%}\\
\cdashlinelr{1-4}
Tower-v2 70B 
& \bf -2.79 & \bf 84.71 & NA \\
\bottomrule
\end{tabular}
\caption{\label{table:metricx-results} Translation quality measured with \textsc{metricX} (M) and \textsc{Comet} (C) on the WMT24 test set.
Win rates against Tower-v2 70B, according to M. 
The bars represent the proportions of \textcolor{tikz_red}{losses}, \textcolor{tikz_gray}{ties}, and \textcolor{tikz_green}{wins}.
Following \citet{kocmi-etal-2024-navigating}, translations with differences in M below 0.122 are considered ties (90\% human accuracy).
}
\end{table}

\subsection{Larger is not necessarily better}
Although Tower-v2 70B outperforms Tower-v2 7B across all language pairs (\cref{table:metricx-results} shows aggregated results), a closer look at its win rates shows it only outperforms the smaller model in $43\%$ of individual examples. This confirms that larger models do not consistently do better on every example, opening the possibility of using smaller models for a subset of examples without compromising overall performance, thus improving efficiency.

\begin{figure}[t]
    \centering
    \includegraphics[width=0.99\linewidth]{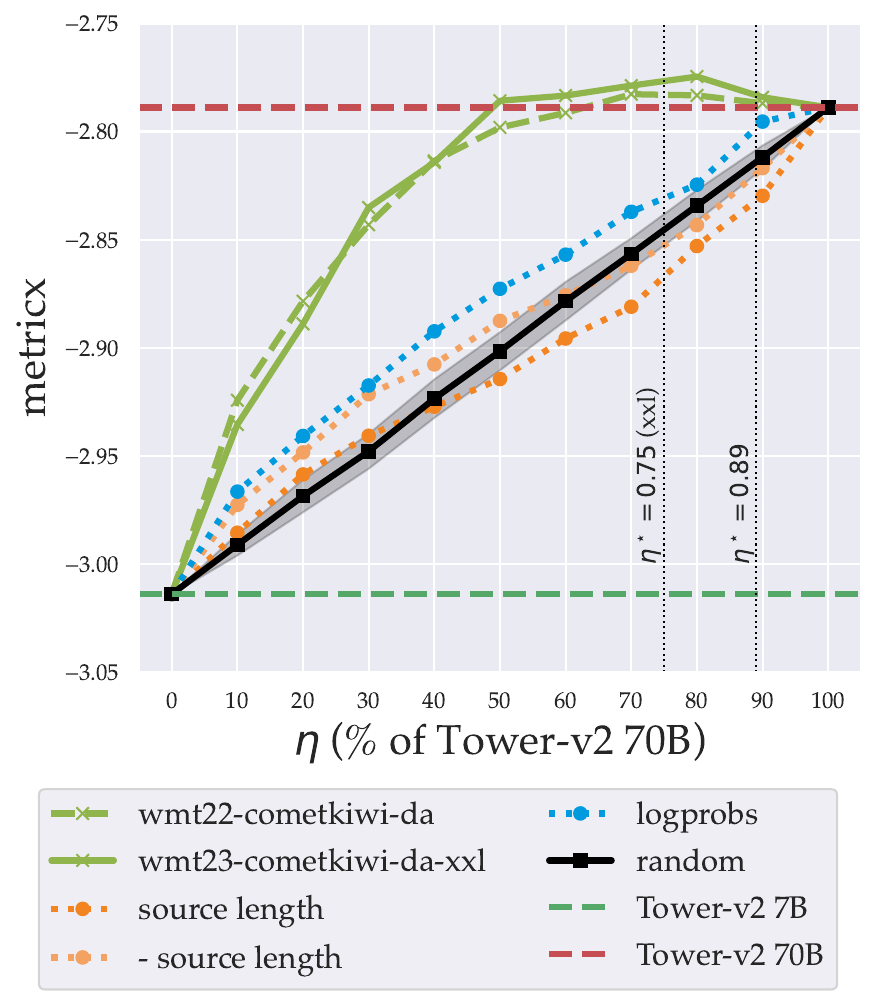}
    \caption{Translation quality of cascading combining Tower-v2 7B and Tower-v2 70B according to \textsc{MetricX}, as the inference computation budget varies. Horizontal lines show the performance of each model alone.}
    \label{fig:qe-is-effective}
\end{figure}

\subsection{QE is an effective deferral rule}
\cref{fig:qe-is-effective} shows the performance of a cascaded system combining Tower-v2 7B and Tower-v2 70B according to \textsc{metricX} under varying inference budgets (results are averaged across language pairs).
Each curve represents a different deferral rule.
As expected, the random baseline fails to identify examples that benefit from larger models, resulting in suboptimal performance.
Source length-based decision rules or using the small model's \texttt{logprobs} perform slightly better or worse than random, suggesting that simple heuristics are inefficient  for deferral.
In contrast, QE-based deferral (our proposal) 
achieves the best overall performance, enabling the cascaded system to match the performance of the large model while invoking it for only $50\%$ to $60\%$ of the examples.
From \cref{eq:eta}, computational parity is reached at $\eta^\star=89\%$ when using \texttt{wmt22-cometkiwi-da} ($N_Q=0.5B$) and $\eta^\star=75\%$ with \texttt{wmt23-cometkiwi-da-xxl} ($N_Q=10.5B$).
Matching Tower 70B's performance at such a small $\eta$ shows that our approach effectively balances efficiency and quality.

\begin{figure}[t]
    \centering
    \includegraphics[width=0.99\linewidth]{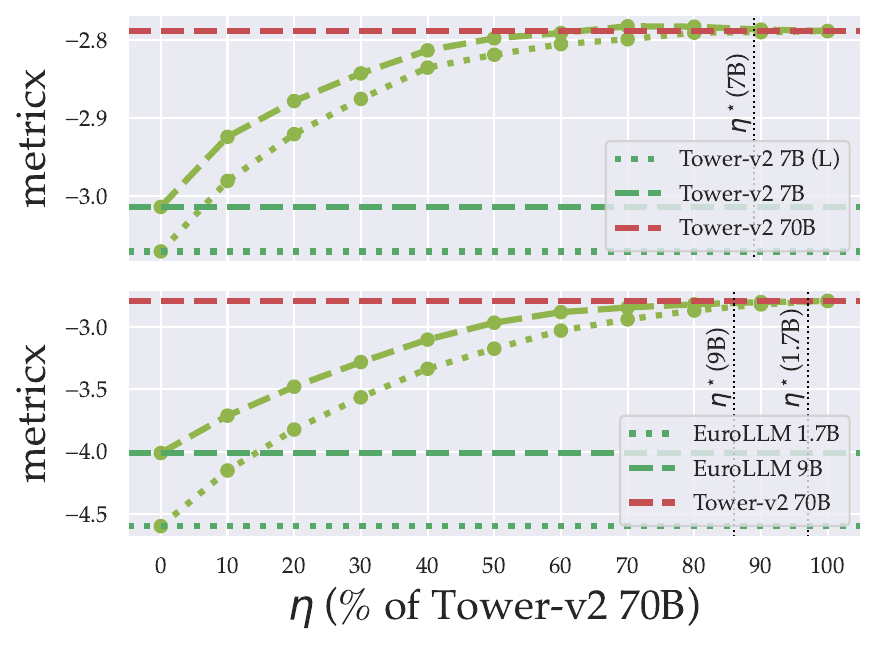}
    \caption{Translation quality of cascaded systems with deferral based on \textcolor{tikz_green}{\texttt{wmt22-cometkiwi-da}}. Large model: {Tower-v2 70B}. Small models: Tower-v2 7B (L), Tower-v2 7B (top); EuroLLM 1.7B, EuroLLM 9B (bottom).}
    \label{fig:llama-vs-mistral}
\end{figure}

\subsection{Cascading works across different setups}

We have shown that QE-based cascading works well across QE models of different sizes (\cref{fig:qe-is-effective}). Here, we study whether it still provides gains when the smaller model is weaker.
We train another version of Tower 7B using Llama-3 instead of Mistral, referred to as \textbf{Tower 7B (L)}, and use two versions of \textbf{EuroLLM} \citep{martins2024eurollm} with 1.7B ($\eta^\star=0.97$) and 9B parameters ($\eta^\star=0.86$).
\cref{fig:llama-vs-mistral} shows that while these models underperform Tower-v2 7B, cascading with Tower 70B remains competitive.
This indicates that QE-based cascading is robust across different generation models, even when both belong to the same family (top) or when the small model is much smaller (bottom).

\begin{figure}[t]
    \centering
    \includegraphics[width=0.99\linewidth]{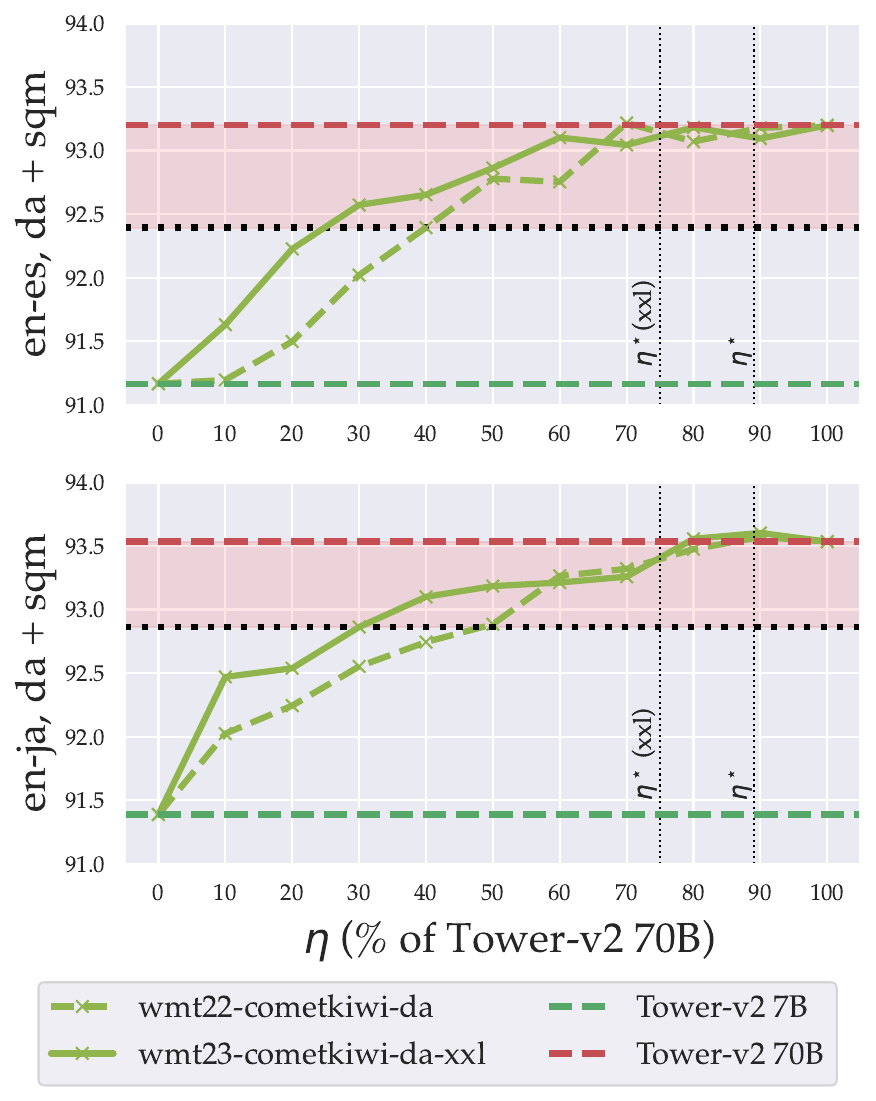}
    \caption{Translation quality of a cascaded system combining Tower-v2 7B and Tower-v2 70B according to human scores (in a scale from 0 to 100), as the inference computation budget varies. Systems in the shaded area are not significantly different from Tower-v2 70B according to the paired-permutation test with $p=0.01$.
    }
    \label{fig:human-eval}
\end{figure}

\section{Human Evaluation}
\label{sec:Human Evaluation}

Since using QE metrics during inference can bias automatic evaluations, we conduct a human study to validate our approach.
We randomly sample 500 source instances and ask human annotators to rate translations from Tower-v2 7B and Tower-v2 70B on a continuous scale from 1 (no overlap in meaning) to 100 (perfect translation). This is done for en-es and en-ja. Further details are in \cref{app:Human Evaluation}.

\cref{fig:human-eval} shows the performance of cascaded systems using QE-based deferral.
We use a paired-permutation test \citep{good2013permutation, zmigrod-etal-2022-exact} to compare the performance of Tower-v2 70B with our systems under varying budgets.
The shaded region shows that our approach achieves performance comparable to Tower-v2 70B while invoking it for only $30\%$ to $50\%$ of the examples,\footnote{Systems within the shaded region are also significantly better than Tower-v2 7B according to the same statistical test.} confirming that it substantially reduces computational costs without compromising translation quality. \cref{App:Deferral based on other QE metrics} provides further evidence using other QE models.

\section{Conclusions and Future Work}

We propose a simple yet effective approach to model cascading for MT using QE metrics for deferral. Our method matches the quality of larger models while requiring them to handle only a subset of examples, significantly reducing computational costs. This is shown through automatic and human evaluations.
The effectiveness of our framework depends on the quality of existing QE models, and improving them can further strengthen our approach (\cref{App:Oracles}).

\section{Limitations}
We highlight three main limitations of our work.
First, we focus on a two-stage cascade, where examples are handled by a small model or deferred to a larger one. Extending this to a multistage setup with more than two models could further improve efficiency but also add complexity.
Second, our study is limited to machine translation. QE-based deferral works particularly well in MT due to the availability of high-quality human-labeled data for training QE models. Extending this approach to other tasks where such data is scarce is not straightforward.
Finally, our method assumes the smaller model is reasonably competitive with the larger one, which is a fair assumption for MT, as shown in our experiments. If the gap in win rates is too large, cascading offers little benefit, as most examples would require deferral.

\section*{Acknowledgments}
We thank Duarte Alves, Patrick Fernandes, Emmanouil Zaranis, and the SARDINE lab team for helpful discussions. This work was supported by EU's Horizon Europe Research and Innovation Actions (UTTER, contract 101070631), by the project DECOLLAGE (ERC-2022-CoG 101088763), by the Portuguese Recovery and Resilience Plan through project C645008882-00000055 (Center for Responsible AI), and by FCT/MECI through national funds and when applicable co-funded EU funds under UID/50008: Instituto de Telecomunicações. 

\bibliography{custom}

\newpage
\appendix

\section{Experimental Details}
\label{app:Implementation details}
Through the paper, we experiment with the following generation models:
\begin{itemize}
    \item \textbf{Tower-v2 70B} \citep{rei-etal-2024-tower}: An improved iteration of Tower \cite{alves2024tower}, obtained by continued pertaining Llama-3 \citep{llama3modelcard} on a multilingual dataset with billions of tokens, followed by supervised finentuning for translation-related tasks. It has 70B parameters. Compared to the first iteration of Tower, this model is better at paragraph and document-level translation and supports more language (15, instead of 10), including all the languages in the WMT24 test sets. Combined with quality-aware decoding \citep{fernandes-etal-2022-quality}, this is the winning submission of the WMT24 general translation shared task \citep{kocmi-etal-2024-findings}.
    \item \textbf{Tower-v2 7B} \citep{rei-etal-2024-tower}: A smaller version of Tower-v2 70B based on Mistral \citep{jiang2023mistral7b}.
    \item \textbf{Tower-v2 7B (Llama-3)}: We follow the recipe described above to train a smaller version of Tower-v2 70B based on LLama-3. This model slightly underperforms its Mistral counterpart.
    \item \textbf{EuroLLM Instruct (9B and 1.7B)} \citep{martins2024eurollm}: EuroLLM models are open-weight multilingual models trained on 4 trillion tokens covering all European Union and many other relevant languages across several data sources: web data, parallel data (en-xx and xx-en), and high-quality datasets. The instruction-tuned models are obtained after finetuning the base models on the EuroBlocks dataset, which includes general instruction-following and machine translation tasks.
\end{itemize}

We generate all translations with greedy decoding using vLLM \citep{kwon2023efficient} for faster inference. \cref{table:metricx-results-full} shows the performance of these models on the WMT24 test sets \citep{kocmi-etal-2024-findings},\footnote{Publicly available for research purposes at \url{https://www2.statmt.org/wmt24/translation-task.html}.} according to \textsc{MetricX} and \textsc{Comet} (results are averaged across all language pairs), along with their win rates against Tower-v2 70B.\footnote{Following \citet{kocmi-etal-2024-navigating}, translations with differences in \textsc{MetricX} below 0.122 are considered ties when comparing two systems (90\% human accuracy). We use the same threshold for detecting ties at the segment level.} Our use of datasets and models aligns with their intended purposes as defined by the licenses.

\section{Quality-Aware Decoding}
\label{App:Quality-Aware Decoding}

There is a large body of work on \textbf{reranking for language generation}, where we start by generating multiple hypotheses with a language model, and then use a reranker to select the best one \citep{farinhas2024rerankinglawslanguagegeneration}.
For machine translation, an example is quality-aware decoding \citep{fernandes-etal-2022-quality, freitag-etal-2022-high}. The simplest/cheapest approach is QE reranking, where we first generate multiple translation hypotheses and then rerank them using a quality estimation model. This strategy is often used to reduce the propensity of language models to hallucinate or generate critical errors \citep{guerreiro-etal-2023-looking, farinhas-etal-2023-empirical}.
While our approach is conceptually different---designed with efficiency in mind, whereas QE reranking is often computationally expensive---it is nonetheless valuable to compare its performance against QE reranking based on hypotheses generated by the small model.

\begin{table}[t]
\small
\centering
\setlength\tabcolsep{4pt}
\begin{tabular}{l ccc c}
\toprule
& M $\uparrow$ & C $\uparrow$ & Win rate \\
\midrule
Tower-v2 7B & -3.01 & 83.94 & 
\scriptsize{\textcolor{tikz_red}{43\%}}
\begin{tikzpicture}
\draw [fill=tikz_red] (0,0) rectangle (0.43*1.4, 0.2);
\draw [fill=tikz_gray] (0.43*1.4,0) rectangle (0.68*1.4, 0.2); 
\draw [fill=tikz_green] (0.68*1.4,0) rectangle (1*1.4, 0.2); 
\end{tikzpicture} \scriptsize{\color{tikz_green}32\%} \\
Tower-v2 7B (L) & -3.07 & 83.73 & 
\scriptsize{\textcolor{tikz_red}{45\%}}
\begin{tikzpicture}
\draw [fill=tikz_red] (0,0) rectangle (0.45*1.4, 0.2);
\draw [fill=tikz_gray] (0.45*1.4,0) rectangle (0.68*1.4, 0.2);
\draw [fill=tikz_green] (0.68*1.4,0) rectangle (1*1.4, 0.2); 
\end{tikzpicture} \scriptsize{\color{tikz_green}32\%} \\
EuroLLM 9B & -4.01 & 80.56 &
\scriptsize{\textcolor{tikz_red}{52\%}}
\begin{tikzpicture}
\draw [fill=tikz_red] (0,0) rectangle (0.52*1.4, 0.2);
\draw [fill=tikz_gray] (0.52*1.4,0) rectangle (0.72*1.4, 0.2);
\draw [fill=tikz_green] (0.72*1.4,0) rectangle (1*1.4, 0.2); 
\end{tikzpicture} \scriptsize{\color{tikz_green}28\%} \\
EuroLLM 1.7B & -4.60 & 77.42 &
\scriptsize{\textcolor{tikz_red}{66\%}}
\begin{tikzpicture}
\draw [fill=tikz_red] (0,0) rectangle (0.66*1.4, 0.2); 
\draw [fill=tikz_gray] (0.66*1.4,0) rectangle (0.80*1.4, 0.2); 
\draw [fill=tikz_green] (0.80*1.4,0) rectangle (1*1.4, 0.2); 
\end{tikzpicture} \scriptsize{\color{tikz_green}20\%} \\
\cdashlinelr{1-4}
Tower-v2 70B & \bf -2.79 & \bf 84.71 & NA \\
\bottomrule
\end{tabular}
\caption{\label{table:metricx-results-full} Translation quality measured with \textsc{MetricX} (M) and \textsc{Comet} (C) on the WMT24 test set.
Win rates against Tower-v2 70B, according to M. 
The bars represent the proportions of \textcolor{tikz_red}{losses}, \textcolor{tikz_gray}{ties}, and \textcolor{tikz_green}{wins}.
}
\end{table}

\paragraph{Computational efficiency.}
Following the discussion in \cref{sec:An Embarrassingly Easy Approach for Machine Translation}, the number of FLOPS required for inference with a large model on a batch of $B$ examples is given by:
\begin{equation}
    2BDN_L,
\end{equation}
where $N_L$ represents the number of model parameters and $D$ is the number of generated tokens.
In this section, we assume that our goal is to \textbf{reduce the computational cost by $\boldsymbol{(1-X)}\boldsymbol{\%}$}, meaning that we operate under a computational budget of:
\begin{equation}
    X\cdot2BDN_L.
\end{equation}
The number of FLOPs required to run inference with our cascaded approach is given by:
\begin{equation}
    2BD (N_S+N_{QE}+\eta N_L),
\end{equation}
which leads to the following expression for $X$:
\begin{equation}
    X = \eta + \frac{N_S+N_{QE}}{N_L}.
\end{equation}
For QE reranking, the computational cost is:
\begin{equation}
    2BDK(N_S + N_{QE}),
\end{equation}
where $K$ is the number of generated hypotheses. This yields:
\begin{equation}
    X = K \cdot \left(\frac{N_S+N_{QE}}{N_L}\right). \label{eq:X-qe-reranking}
\end{equation}
These expressions allow us to obtain the values of $\eta$ for which our cascaded approach incurs the same computational cost as QE reranking with $K$ hypotheses:
\begin{equation}
     \eta = (K-1) \cdot \left(\frac{N_S+N_{QE}}{N_L}\right).
\end{equation}

\begin{figure}[t]
    \centering
    \includegraphics[width=0.99\linewidth]{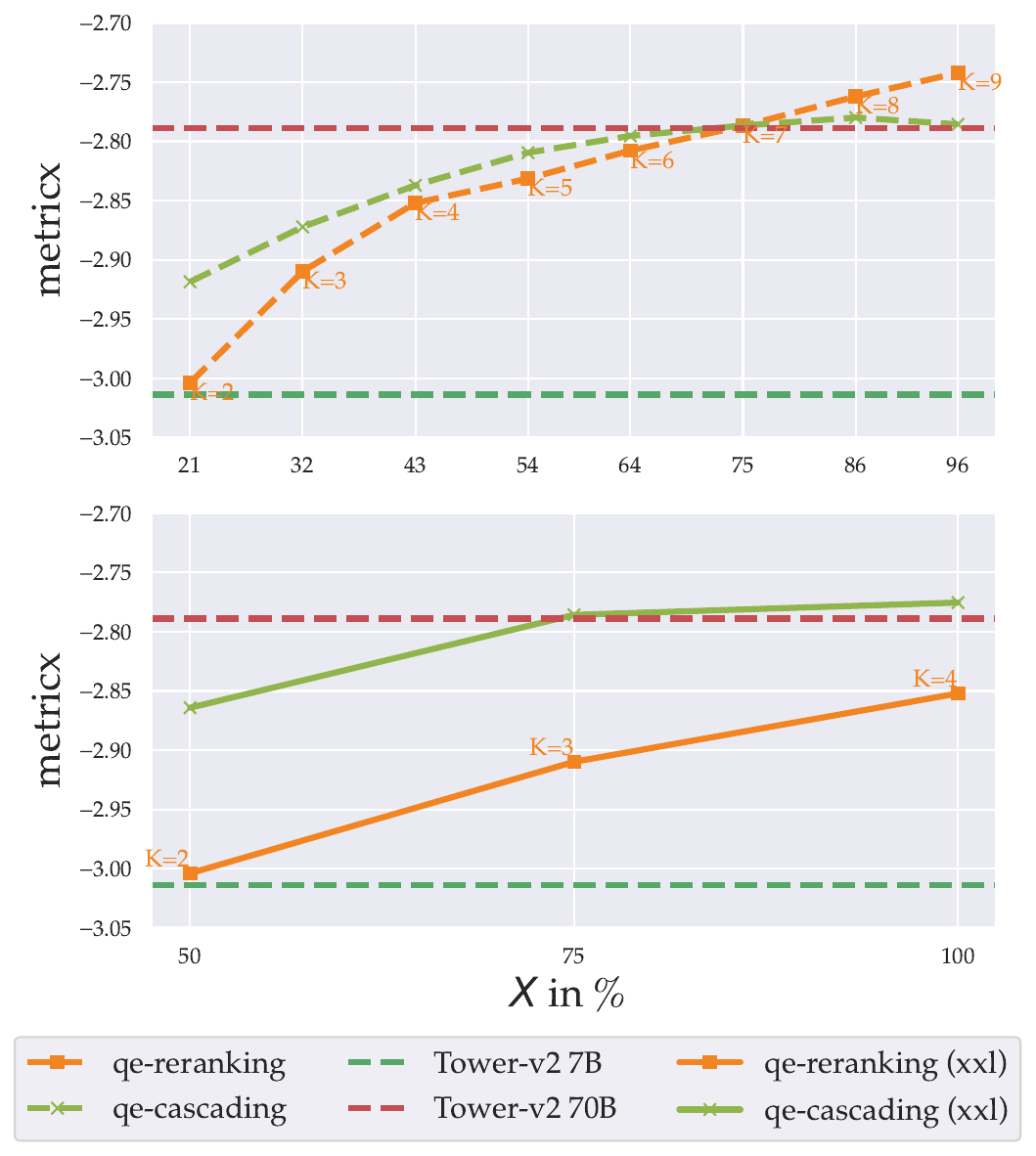}
    \caption{Translation quality of a cascaded system combining Tower-v2 7B and Tower-v2 70B (in \textcolor{tikz_green}{green}) \textit{v.s.} QE reranking with hypotheses generated by Tower-v2 7B (in \textcolor{tikz_orange}{orange}), measured with \textsc{MetricX}, as $X$ varies. Horizontal lines show the performance of the smaller and larger models alone.
    }
    \label{fig:comparision-qe-reranking}
\end{figure}

\paragraph{Experiments and discussion.}
We generate up to $9$ hypotheses with Tower-v2 7B using $\epsilon$-sampling with $\epsilon=0.02$ \citep{freitag-etal-2023-epsilon}.\footnote{For our setup, according to \cref{eq:X-qe-reranking}, the number of FLOPs required for QE reranking with more than $9$ hypotheses already exceeds the budget of $2BDN_L$ if we use \texttt{wmt22-cometkiwi-da}. When using \texttt{wmt23-cometkiwi-da-xxl}, computational parity is achieved with $K=4$.}
\cref{fig:comparision-qe-reranking} illustrates the trade-off between computational efficiency and translation quality (measured with \textsc{metricX}) for a 
cascaded approach with QE-based deferral \textit{versus} QE reranking.
As expected, quality improves as the computational budget increases for both methods.
While QE reranking is an effective way to improve translation quality when generating multiple hypotheses is feasible, our cascaded approach achieves better quality at lower computation costs, making it a more efficient alternative when computational efficiency is a priority.

\section{Human Evaluation}
\label{app:Human Evaluation}

In order to perform human evaluation, we recruited professional translators who were native speakers of the target language on the freelancing site Upwork.\footnote{\url{https://upwork.com}}
We followed a DA+SQM (direct assessment + scalar
quality metric) source contrastive evaluation \citep{kocmi-etal-2022-findings} using Appraise \citep{federmann-2018-appraise}.
We sampled 500 source instances from the WMT24 test set for en-ja and en-es and asked one translator per language pair to read two alternative translations for each source and evaluate them on a continuous scale from 0 to 100.
The scale featured seven labeled tick marks (from 0 to 6) indicating different quality labels combining \textit{accuracy} and \textit{grammatical correctness}.
Translators could further adjust their scores to reflect preferences or assign the same score to translations of similar quality.
They were paid a market rate of around 20 USD per hour, and completing the task took approximately 12 to 14 hours for each language pair.

\begin{figure}[b]
    \centering
    \includegraphics[width=0.99\linewidth]{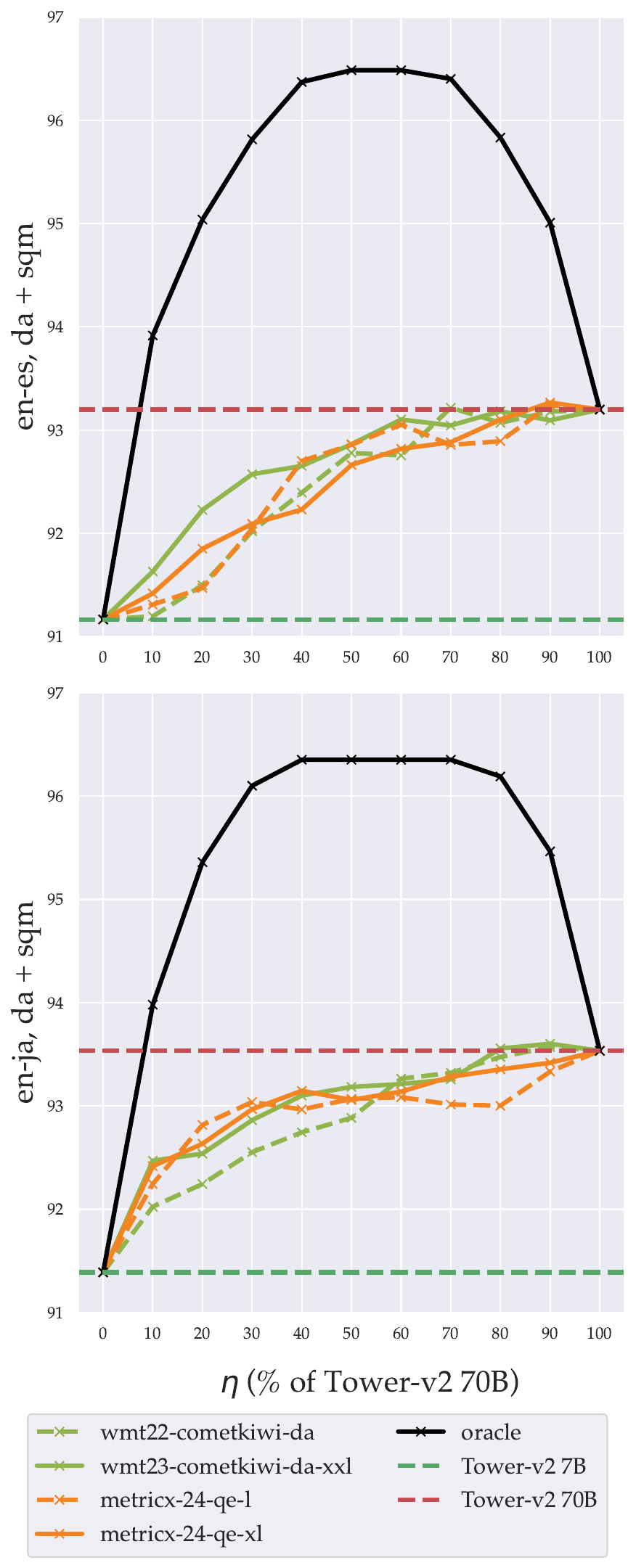}
    \caption{Translation quality of a cascaded system combining Tower-v2 7B and Tower-v2 70B according to human scores (in a scale from 0 to 100), as the inference computation budget varies. Deferral is based on different QE models (\textcolor{tikz_green}{green} and \textcolor{tikz_orange}{orange} curves). The black curve shows the oracle selection.
    }
    \label{fig:human-eval-metricx}
\end{figure}

\subsection{Deferral based on other QE metrics}
\label{App:Deferral based on other QE metrics}

We have seen that QE-based cascading works well with \textsc{CometKiwi} models of different sizes (\cref{fig:human-eval}).
Here, we show that this is also the case when using two reference-free versions of \textsc{MetricX} \citep{juraska-etal-2024-metricx}: \texttt{metricx-24-hybrid-large-v2p6} and \texttt{metricx-24-hybrid-xl-v2p6} (\cref{fig:human-eval-metricx}, \textcolor{tikz_orange}{orange} curves).

\subsection{Oracle selection}
\label{App:Oracles}
The effectiveness of our framework depends on the quality of existing QE models, and improving them can further strengthen our approach. To access the performance ceiling of cascading, we report results with oracle deferral, \textit{i.e.}, a deferral strategy that maximizes translation quality according to humans (\cref{fig:human-eval-metricx}, black curves).\footnote{Oracle performance goes down after reaching a \textit{plateau} due to our budget-constrained approach, which enforces deferral for a fixed percentage of examples.} The high oracle values indicate significant potential for improvement, suggesting that having better QE models could directly boost the effectiveness of our cascaded approach.

\begin{figure*}[t]
    \centering
    \includegraphics[trim={0cm 12cm 0cm 0cm},clip,width=0.99\linewidth]{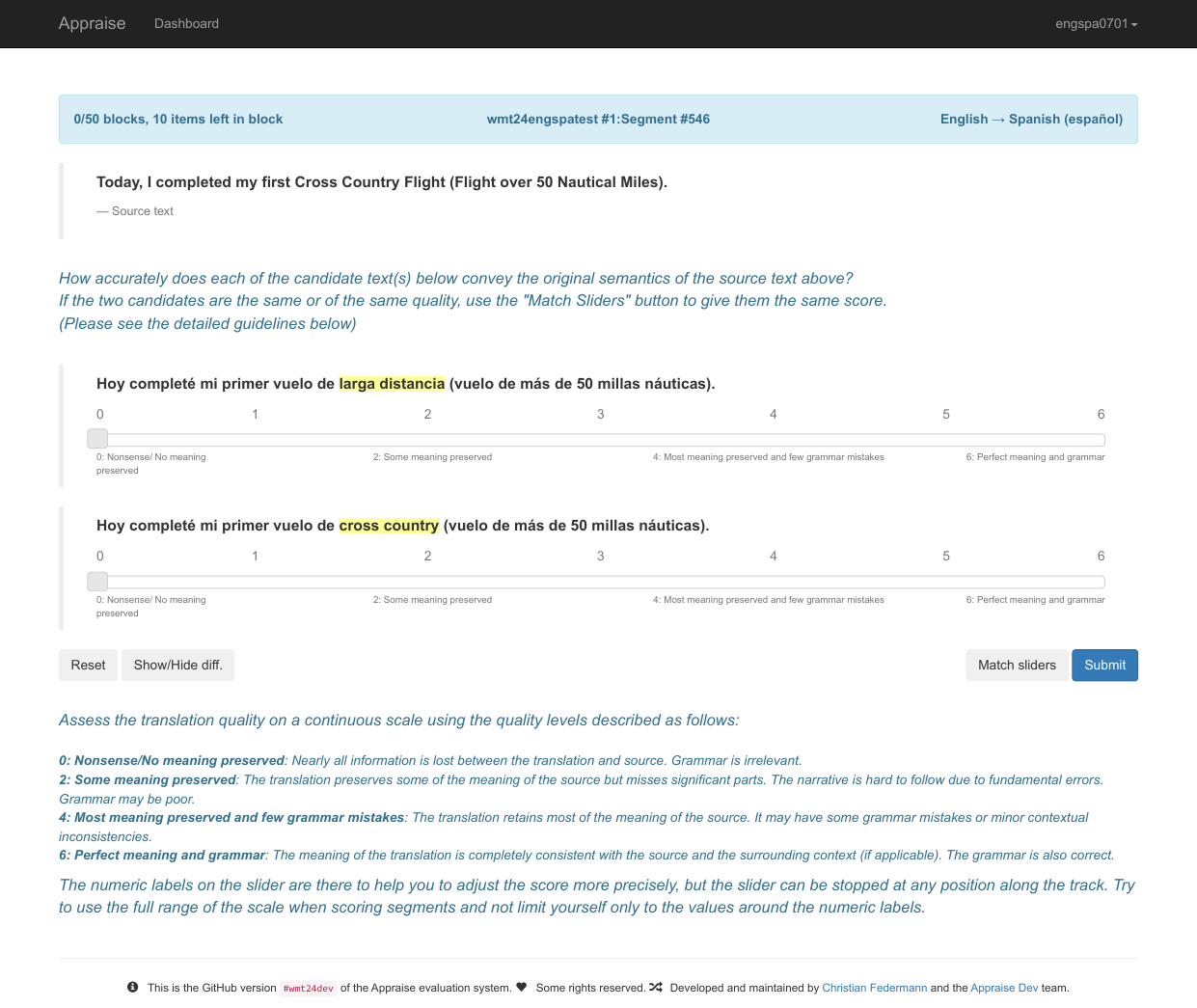}
    \caption{Annotation interface for en-es.}
    \label{fig:annotation-interface-es}
\end{figure*}

\begin{figure*}[t]
    \centering
    \includegraphics[trim={0cm 12cm 0cm 0cm},clip,width=0.99\linewidth]{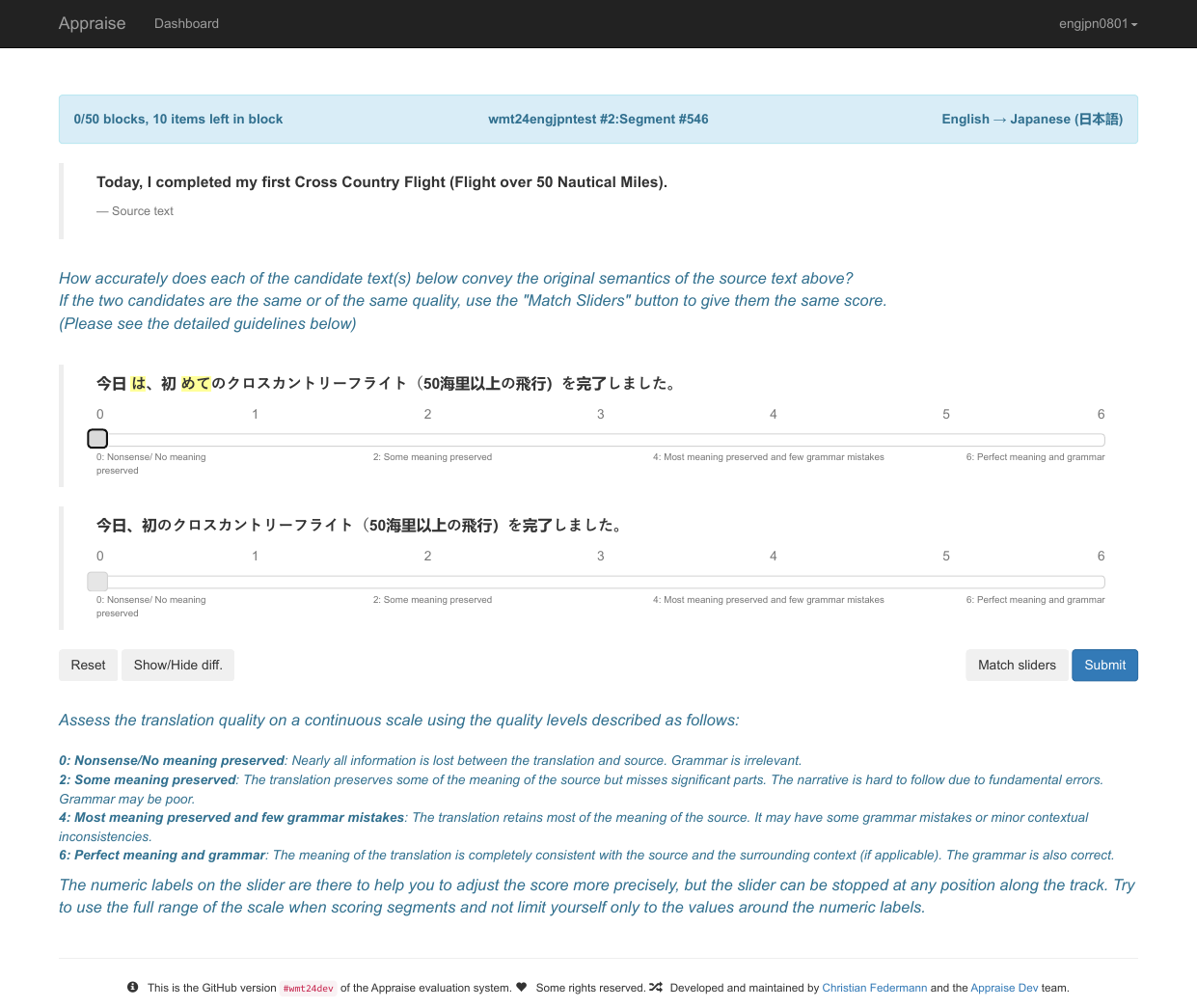}
    \caption{Annotation interface for en-ja.}
    \label{fig:annotation-interface-ja}
\end{figure*}

\subsection{Annotation guidelines}
We share below the annotation guidelines shared with the freelancers. 

\paragraph{Task overview.}
This task involves evaluating two alternative translations of a source text and assigning a rating to each translation based on its overall quality and adherence to the source content. You should consider accuracy, fluency, and overall quality when assessing the different translations.

\paragraph{Annotation scale.}
Each translation should be evaluated on a continuous scale from 0 to 6 with the quality levels described below:
\begin{itemize}
    \item \textbf{6 (perfect meaning and grammar):} The meaning of the translation is completely consistent with the source and the surrounding context, if applicable. The grammar is also correct. 
    \item \textbf{4 (most meaning preserved and few grammar mistakes):} The translation retains most of the meaning of the source. It may have some grammar mistakes or minor contextual inconsistencies. 
    \item \textbf{2 (some meaning preserved):} The translation preserves some of the meaning of the source but misses significant parts. The narrative is hard to follow due to fundamental errors. Grammar may be poor. 
    \item \textbf{0 (nonsense/no meaning preserved):} Nearly all information is lost between the translation and source. Grammar is irrelevant. 
\end{itemize}

\paragraph{Annotation interface.}
\cref{fig:annotation-interface-es,fig:annotation-interface-ja} show the annotation interface.
If two candidates were the same or of the same quality, the annotators were asked to use ``\textbf{match sliders}'' to give them the exact same score. And, they could also use the absolute scale range to show preference between the translations.

\end{document}